%% file: elsarticle-num.tex
\documentclass[times, review, 10pt]{elsarticle}

\usepackage{amssymb}
\usepackage{amsmath}

\usepackage{url}
\usepackage{lineno}
\usepackage{graphicx}
\usepackage{multirow}
\usepackage{rotating}
\usepackage{subcaption}
\usepackage{longtable}
\usepackage{booktabs}
\usepackage[dvipsnames]{xcolor}
\usepackage[normalem]{ulem}

\journal{Pattern Recognition}

\begin{document}

\begin{frontmatter}

\title{Balancing Tails when Comparing Distributions: Comprehensive Equity Index (CEI) with Application to Bias Evaluation in Operational Face Biometrics}

\author[VeridasLabel,BiDALabel]{Imanol Solano\corref{cor1}}
\author[BiDALabel]{Julian Fierrez\corref{cor2}}
\author[BiDALabel]{Aythami Morales}
\author[BiDALabel]{Alejandro Peña}
\author[BiDALabel]{Ruben Tolosana}
\author[VeridasLabel]{Francisco Zamora-Martinez}
\author[VeridasLabel]{Javier San Agustin}

\affiliation[VeridasLabel]{organization={Veridas},
             city={Pamplona},
             country={Spain}}
\affiliation[BiDALabel]{organization={BiDA Lab, UAM},
             city={Madrid},
             country={Spain}}

\cortext[cor1]{Corresponding author: imanolsolano97@gmail.com}
\cortext[cor2]{Corresponding author: julian.fierrez@uam.es}

\input{sections/0_abstract}

\begin{keyword}
Biometrics \sep Face Recognition \sep Fairness \sep Bias
\end{keyword}

\end{frontmatter}


\input{sections/1_introduction}

\input{sections/2_previous_work}

\input{sections/3_methodology}

\input{sections/4_experimental_protocol}

\input{sections/5_results_and_interpretation}

\input{sections/6_conclusions}


\bibliographystyle{elsarticle-num} 
\bibliography{sections/refs}

\end{document}

%% file: sections/0_abstract.tex
\begin{abstract}
Demographic bias in high-performance face recognition (FR) systems often eludes detection by existing metrics, especially with respect to subtle disparities in the tails of the score distribution. We introduce the Comprehensive Equity Index (CEI), a novel metric designed to address this limitation. CEI uniquely analyzes genuine and impostor score distributions separately, enabling a configurable focus on tail probabilities while also considering overall distribution shapes. Our extensive experiments (evaluating state-of-the-art FR systems, intentionally biased models, and diverse datasets) confirm CEI's superior ability to detect nuanced biases where previous methods fall short. Furthermore, we present CEI\textsuperscript{A}, an automated version of the metric that enhances objectivity and simplifies practical application. CEI provides a robust and sensitive tool for operational FR fairness assessment. The proposed methods have been developed particularly for bias evaluation in face biometrics but, in general, they are applicable for comparing statistical distributions in any problem where one is interested in analyzing the distribution tails.
\end{abstract}

%% file: sections/1_introduction.tex
\section{Introduction}
\label{sec:intro}

Artificial Intelligence (AI) now plays an increasingly crucial role in modern life, from personalized treatment recommendations in healthcare~\cite{gomez2021improving}, to adaptive learning platforms in education~\cite{daza2023edbb}. Although AI offers enhanced efficiency and scalability, its growing role in critical processes has intensified concerns about fairness, accountability, and transparency~\cite{2023_SNCS_Human-Centric_Pena}.

Biometric recognition, a key AI application, offers advantages over traditional security methods and is widely used~\cite{jain2011introduction}. Among various traits, face recognition (FR) has become particularly prominent. Advances in deep learning, including novel architectures~\cite{He2015DeepRL} and loss functions~\cite{Deng2018ArcFaceAA}, have significantly improved the performance of FR, leading to its deployment in diverse applications such as border control and mobile authentication.

However, progress in FR is affected by significant ethical concerns, particularly algorithmic discrimination~\cite{serna2020aaai} and demographic biases~\cite{2018_CIARP_BiasFace_Acien,sebastien25}. Biased biometric systems can lead to unfair treatment based on the membership of demographic groups, an issue highlighted by regulatory frameworks such as the GDPR\footnote{\url{https://gdpr-info.eu/}}, the EU AI Act\footnote{\url{https://www.euaiact.com/}}, and new ISO standards for fairness in biometrics\footnote{\url{https://www.iso.org/standard/81223.html}}.

Addressing demographic bias requires effective measurement~\cite{serna25unravel}. Although research has explored bias detection~\cite{2022_SafeAI_IFBiD_Serna}, evaluation~\cite{serna2021insidebias}, and mitigation strategies~\cite{Gong2019JointlyDF,wang2020mitigating,mancera2025pba}, traditional bias measurement often relies on comparing error rates (e.g., False Match Rate, FMR; False Non-Match Rate, FNMR) between demographic groups~\cite{howard2019effect}. Metrics such as the Fairness Discrepancy Rate (FDR)~\cite{deFreitasPereira2020FairnessIB} and the Gini Aggregation Rate for Biometric Equitability (GARBE)~\cite{Howard2022EvaluatingPF} assess fairness at specific operating thresholds. However, evaluating high-performance systems, particularly those with long-tail score distributions, presents unique challenges to existing metrics. Our main contributions are as follows.

\begin{itemize}
\item An evaluation of existing fairness metrics in face recognition, demonstrating their limitations in characterizing bias in high-performance systems with long-tail score distributions, especially with commercial algorithms and state-of-the-art datasets.
\item The proposal of a new metric, the Comprehensive Equity Index (CEI), designed to overcome the drawbacks of metrics such as the Distribution Fairness Index (DFI) by assessing bias across both the tails and the central regions of score distributions in such systems.
\item The proposed methods have been motivated and developed particularly for bias evaluation in face biometrics, but, in general, they are applicable for comparing statistical distributions in any problem where one is interested in analyzing the distribution tails.
\end{itemize}

This paper significantly extends our preliminary work presented in~\cite{2024_cei_icpr}. Key enhancements include: (i) a more robust evaluation with additional models and diverse training datasets; (ii) new experiments involving intentionally biased systems and state-of-the-art high-performance models to test metric sensitivity and applicability; and (iii) a novel algorithmic approach for automatically optimizing parameters in fairness metric calculations.

By extending our previous work, we aim to provide a comprehensive evaluation of fairness metrics and offer novel tools to measure and mitigate demographic bias in biometric systems. The remainder of this paper is  organized as follows. Section~\ref{sec:previous_work} reviews related work. Section~\ref{sec:methodology} details our proposed methodology. Section~\ref{sec:experimental_framework} describes the experimental setup. Finally, Section~\ref{sec:experiments_and_interpretation} presents and discusses our experimental results and the analysis of the proposed metrics.

%% file: sections/2_previous_work.tex
\section{Previous Work}
\label{sec:previous_work}

A face recognition system is usually defined as a feature extractor \( \mathbf{w}^F \) trained to extract invariant identity-discriminative feature vectors \( \mathbf{x} = f(\mathbf{I} \mid \mathbf{w}^F) \), called embeddings, where \( \mathbf{I} \) is a face image. Formally, we can define a face recognition model as \( f: \mathbf{I} \rightarrow \mathbb{R}^d \). Feature vectors are then used to perform either i) verification or ii) identification tasks. We will focus on the first one, consisting of \( 1:1 \) comparisons of different biometric samples, where the objective is to determine whether two samples -typically face images - belong to the same identity. This is done through a \textit{similarity} comparison, producing a score \( s \in [0, 1] \) that reflects how similar the samples are: the closer the score to 1, the more likely they represent the same individual.
A final decision is made using a threshold \( \tau \): if \( s \geq \tau \), the pair is classified as a \textit{genuine} match (same identity); otherwise, it is considered a \textit{impostor} match (or non-match) (different identities). The similarity between two representations \( (\mathbf{x}_n, \mathbf{x}_m) \) is typically computed using a distance metric like cosine similarity, which can be normalized~\cite{Fierrez-Aguilar2005_ScoreNormalization} to yield a similarity score \( s(\mathbf{x}_n, \mathbf{x}_m) \in [0, 1] \). When evaluating the performance of these systems, a dataset of biometric samples consisting of \( N \) face images \( \mathbf{I} \) is considered. Each image corresponds to a subject belonging to a demographic group \( d_i \), determined by attributes such as gender, ethnicity, or age. With this, we can define \( \mathcal{D} \) as the set of \( K \) disjoint demographic groups (each subject belongs to only one group).

In traditional Machine Learning (ML), fairness is often associated with unwanted biases~\cite{mehrabi2021survey,2019_naturemachineintelligence_global_landscape_AI_ethics_guide}: the tendency of a system to develop some kind of favoritism to groups (or individuals) with concrete protected attributes such as ethnicity, age, or sex. In the literature, these preferences are usually related to differences in the model output, where performances are measured as a certain decision of the model (i.e., Demographic Parity~\cite{zafar2017fairness}), or even as the True Positive Rate (i.e., Equality of Odds, and Equality of Opportunity~\cite{hardt2016equality}). These metrics are designed to measure the differences between pairs of demographic groups, where one of the classes is considered preferable (the \textit{positive} class). This is commonly observed in biometrics research~\cite{straga24kvc}, and specially in the face recognition literature~\cite{deFreitasPereira2020FairnessIB, kotwal2023fairness, 2022_AI_SensitiveLoss_IS}. Most existing approaches apply this by considering error rates, that is, the False Match Rate (FMR)/False Non-Match Rate (FNMR), as the basis for these performance measurements, as in~\cite{deFreitasPereira2020FairnessIB}. The main intuition behind these proposals is the same: the "performance" of a system should be the same across different groups, or more formally: $\textrm{Performance}(\mathbf{w}^F|d_i) \approx \textrm{Performance}(\mathbf{w}^F|d_j)\,\, \forall d_i, d_j \in \mathcal{D}$, where $d_i$ is the $i$ demographic group.

In operational environments, these metrics, known as outcome-based metrics, are designed to provide a single score summarizing differences for all demographic groups by using a posterior aggregation, such as those proposed in ISO / IEC FDIS 19795-10~\cite{iso19795-10}, which are also highlighted in the latest NIST FRVT report on the measurement of demographic differentials~\cite{grother2022face}. All these metrics have shown really good behavior when evaluating commercial face recognition systems, usually working under operational use cases. The main disadvantage of these outcome-based metrics is their intrinsic relationship to specific operational points (thresholds), as noted in~\cite{kotwal2023fairness}. Additionally, when computing these metrics, very low error rates are required, which usually implies that the differences will be measured over a few samples in the dataset. A recent study by Terhörst et al.~\cite{terhorst2022comprehensive} pointed out that often measured biases can be caused by features of the test images related to image quality, meaning that in the case of differential outcome-based metric, scores may indicate that the model is demographically bias when in reality those few samples used to compute the bias score do not have enough quality, and the measured differences are not really due to demographic bias.

The other known approach is the so-called differential performance, which considers differences in the entire distributions of scores $z=p(s|\mathbf{w}^F)$ (where $p$ denotes probability) to represent the performance of the model, as introduced in~\cite{kotwal2023fairness}. This approach uses the whole distribution, overcoming the problem that outcome-based metrics have related to the use of a few samples. Additionally, there is no need to fix a concrete threshold, because the information on modes biases over the whole curve and not only specific operational points. However, we observed that this metric was not able to capture differences when used in operational use cases, where systems achieve really high performances, thus provoking the differences to be found in a few samples in the extreme distribution tails.

%% file: sections/3_methodology.tex
\section{Methodology: Fairness in Operational Biometric Systems}\label{sec:methodology}

This section defines the proposed methodology. We will first review existing bias metrics that operate on these output similarity scores, which can generally be classified into two main categories: those representing differential outcomes and those representing differential performance. Following this overview, we will introduce our proposed metric, which is a differential performance approach. Our proposal tries to overcome some problems detected in previous differential performance metrics that were not able to reveal significant demographic disparities, even when existing. We will then detail the new improvements and extended functionalities we have incorporated into our metric, aiming to automatize some of the parameters that this metric previously had. A summary of all metrics discussed in this section, including both existing approaches and our proposal, is provided in Table~\ref{tab:metrics_summary}.

\begin{table}[!t] 
    \centering
    \caption{Summary of characteristics for discussed fairness metrics in face recognition.}
    \label{tab:metrics_summary}
    \resizebox{\textwidth}{!}{
    \begin{tabular}{@{}llccll@{}}
        \toprule
        \textbf{Metric Name} & \textbf{Type} & \textbf{Variant(s)} & \textbf{Distribution Focus} & \textbf{Range} & \textbf{Interpretation (Fairness)} \\
        \midrule
        \multirow{2}{*}{Inequity (IN)} & \multirow{2}{*}{Outcome} & IN\textsubscript{FMR} & Impostor & \multirow{2}{*}{$[1, \infty)$} & Fair: $1$ \\
        & & IN\textsubscript{FNMR} & Genuine & & Unfair: $>1$ \\
        \midrule
        \multirow{2}{*}{GARBE} & \multirow{2}{*}{Outcome} & GARBE\textsubscript{FMR} & Impostor & \multirow{2}{*}{$[0, 1]$} & Fair: $0$ \\
        & & GARBE\textsubscript{FNMR} & Genuine & & Unfair: $>0$ \\
        \midrule
        \multirow{2}{*}{Distribution Fairness Index (DFI)} & \multirow{2}{*}{Performance} & DFI\textsubscript{N} (Normal) & \multirow{2}{*}{Combined} & \multirow{2}{*}{$[0, 1]$} & Fair: $1$ \\
        & & DFI\textsubscript{E} (Extreme) & & & Unfair: $0$ \\
        \midrule
        \multirow{2}{*}{Comprehensive Equity Index (CEI)} & \multirow{2}{*}{Performance} & CEI\textsubscript{N} (Normal) & \multirow{2}{*}{Genuine \& Impostor} & \multirow{2}{*}{$[0, 1]$} & Fair: $1$ \\
        & & CEI\textsubscript{E} (Extreme) & & & Unfair: $0$\\
        \bottomrule
    \end{tabular}
    }
\end{table}

\subsection{Existing Fairness Metrics}
\label{subsec:metrics}

When evaluating fairness in face recognition systems based on their output similarity scores, existing metrics can be broadly categorized into two main types: differential outcome metrics and differential performance metrics. Differential outcome metrics typically measure fairness by quantifying disparities in classification error rates (e.g., False Match Rate or False Non-Match Rate) across different demographic groups at specific operational thresholds. In contrast, differential performance metrics aim to assess fairness by comparing the overall distributions of similarity scores between groups, often independent of a single operational point.

\subsubsection{Differential Outcome Metrics}
\label{susubsec:differential_outcome_metrics}

Several differential outcome metrics have been highlighted for their utility in assessing demographic fairness. In particular, the National Institute of Standards and Technology (NIST) Face Recognition Vendor Test (FRVT) reports on demographic differences~\cite{grother2022face} discuss metrics such as the inequality metric and the GARBE metric.

The Inequity metric, in its updated version, computes the ratio between the maximum error rate (either FMR or FNMR) observed for any demographic group and the geometric mean of that error rate across all groups. This formulation is considered more robust than previous versions that used the minimum error rate in the denominator. The modified Inequity metrics for FMR and FNMR are defined as follows:

\begin{equation}
    \textrm{IN}_\mathrm{FMR} = \frac{\max_{d_i}\mathrm{FMR}(\tau)}{\mathrm{FMR}_\mathrm{geom}}\label{eq:inequity_fmr}
\end{equation}

\begin{equation}
    \textrm{IN}_\mathrm{FNMR} = \frac{\max_{d_i}\mathrm{FNMR}(\tau)}{\mathrm{FNMR}_\mathrm{geom}}\label{eq:inequity_fnmr}
\end{equation}

\noindent High values for the Inequity metric indicate greater unfairness.

The NIST report also proposes the GARBE metric~\cite{howard2019effect}, inspired by the Gini coefficient used in economics to measure income disparity. GARBE quantifies the dispersion of error rates across demographic groups for both FMR and FNMR:

\begin{equation}
    \textrm{GARBE}_\mathrm{FMR} = \frac{\sum_{i}\sum_{j}|\mathrm{FMR}_{d_i}(\tau) - \mathrm{FMR}_{d_j}(\tau)|}{2K^2\,\mathrm{FMR}_\mathrm{arith}}
\end{equation}

\begin{equation}
    \textrm{GARBE}_\mathrm{FNMR} = \frac{\sum_{i}\sum_{j}|\mathrm{FNMR}_{d_i}(\tau) - \mathrm{FNMR}_{d_j}(\tau)|}{2K^2\,\mathrm{FNMR}_\mathrm{arith}}
\end{equation}

\noindent where $\mathrm{FMR}_\mathrm{arith}$ and $\mathrm{FNMR}_\mathrm{arith}$ represent the arithmetic means of the respective error rates for the demographic groups $K$ considered. The GARBE metric yields values in the interval $[0, 1]$, where higher values signify greater unfairness.

For both Inequity and GARBE metrics, the selection of an operational point, $\tau$, is necessary to compute the FMR and FNMR values for each group. For example, NIST evaluations often fix the operational point such that the overall FMR of the system is $0.0003$. By calculating these metrics separately for FMR and FNMR, we can gain insight into whether a model exhibits more bias regarding false acceptances (FMR) or false rejections (FNMR).

\subsubsection{Differential Performance Metrics}
\label{susubsec:differential_performace_metrics}

Although differential outcome metrics provide valuable information on specific thresholds, the community has also explored differential performance metrics that consider the entire score distribution. Kotwal and Marcel~\cite{kotwal2023fairness} have contributed significantly to this area, arguing that much of the prior attention had been on differential outcome metrics. They proposed a differential performance metric based on the distances between the similarity score distributions ($z$) of different demographic groups. A key advantage of this approach is its independence from any specific operational point, thereby measuring the fairness characteristics of the overall system across its potential range of operation.

The metric introduced by Kotwal and Marcel, known as the Distribution Fairness Index (DFI), utilizes the Kullback-Leibler (KL) divergence to quantify the distance between the score distributions of each demographic group and a reference mean distribution. DFI values range from $0$ to $1$, where a score closer to $1$ indicates a more fair model. Following the notation in~\cite{kotwal2023fairness}, the mean distribution $z_{D_\mathrm{mean}}$ is defined as:

\begin{equation}\label{eq:z_mean}
    z_{D_\mathrm{mean}}=\frac{1}{K} \sum_{i=1}^K z_{D_i}
\end{equation}

\noindent where $z_{D_{i}}$ is the combined score distribution (genuine plus impostor) for the demographic group $d_i$, normalized such that its area is one. The baseline DFI formulation (Normal, denoted DFI\textsubscript{N}), based on the average dissimilarity from the mean distribution, is:

\begin{equation}
    \textrm{DFI}_\mathrm{N}=1-\frac{1}{K\log_2K}\sum_{i=1}^K S_i
    \label{eq:dfi_standard}
\end{equation}

\noindent where $S_i$ is the KL divergence between $z_{D_{i}}$ and $z_{D_\mathrm{{mean}}}$.

To capture the worst-case fairness scenario, an alternative formulation, DFI\textsubscript{E} (Extreme), considers only the demographic group whose distribution diverges most from the mean:

\begin{equation}
    \textrm{DFI}_\mathrm{E}=1-\frac{1}{\log_2K}\max(S_i)
    \label{eq:dfi_extreme}
\end{equation}

\subsection{Proposed Fairness Metric: Comprehensive Equity Index (CEI)}
\label{subsec:CEI_def}

In this section, we introduce the Comprehensive Equity Index (CEI), a fairness metric that extends the work of Kotwal and Marcel~\cite{kotwal2023fairness}. Our proposal aims to retain the advantages of performance-based metrics while incorporating the error-focused perspective characteristic of differential outcome metrics. This balanced approach is intended not only to quantify a model's bias but also to consider the system's operational competitiveness, an important aspect, particularly for high-performance recognition systems that exhibit very low error rates.

Our observations from the evaluation of high performance models (such as those submitted to NIST FRVT) with the DFI metric on datasets such as RFW~\cite{Wang2018RacialFI} or BUPT-B~\cite{wang2020mitigating} indicated that DFI did not consistently capture error rates associated with demographic biases. We hypothesize two primary reasons for this: first, because DFI utilizes the entire score distribution, the tail regions (where errors typically manifest) may have limited influence on the final computation; second, DFI treats genuine and impostor distributions as a single combined entity, potentially obscuring biases specific to either distribution. In contrast, differential outcome metrics such as GARBE~\cite{howard2019effect} or Inequity~\cite{grother2022face} can detect these tail-specific biases due to their reliance on an operational point, which inherently focuses the evaluation on the tails. However, assessing fairness at a single operational point has drawbacks: It may not capture the systemic bias embedded within the broader behavior of the biometric system, and observed outcome differences could arise from factors other than demographic attributes, such as variations in image quality (e.g. resolution, brightness, pose).

The CEI is designed to address these limitations. Our objective is to provide a metric that is threshold-agnostic, capable of assessing bias in genuine and impostor distributions independently, and gives appropriate consideration to the distribution tails where errors occur. To achieve this, for each demographic group, the CEI first divides each score distribution (i.e., genuine or impostor) into two segments based on a specified percentile, $P_{split}$. This percentile corresponds to a score threshold $s$, which separates the tail of the distribution from its main body (hereafter referred to as the center). The rationale is to create distinct components that can be weighted differently in the fairness calculation.

Once a distribution $z_{D_i}$ for a demographic group $d_i$ is split into its tail part ($z_{D_i}^t$) and the center part ($z_{D_i}^c$), we compute a dissimilarity score, $S^{\prime}_{i}$, between the parts of the distribution of this group and the corresponding parts of a mean distribution (calculated according to Eq.~\ref{eq:z_mean} for each part). This score is defined as follows:

\begin{equation}
    S^{\prime}_{i}(P_{split}) = w_{t} \cdot D_\mathrm{KL}(z_{D_i}^t \,||\, z_{D_\mathrm{mean}}^t) + w_{c} \cdot D_\mathrm{KL}(z_{D_i}^c \,||\, z_{D_\mathrm{mean}}^c)
    \label{eq:s_prime}
\end{equation}

\noindent where $D_\mathrm{KL}$ denotes the Kullback-Leibler divergence, $z_{D_\mathrm{mean}}^t$ and $z_{D_\mathrm{mean}}^c$ are the tail and center parts of the mean distribution (derived from applying the same logic $P_{split}$ to $z_{D_\mathrm{mean}}$), and $w_{t}$ and $w_{c}$ are weights that control the relative importance of the tail and center dissimilarities, respectively, with $w_t + w_c = 1$. After computing $S^{\prime}_{i}(P_{split})$ for each demographic group, the CEI is calculated similarly to the DFI, producing variants Normal (CEI\textsubscript{N}) and Extreme (CEI\textsubscript{E}) variants:

\begin{equation}
    \textrm{CEI}_\mathrm{N} (P_{split}) = 1-\frac{1}{K\log_2K}\sum_{i=1}^K S^{\prime}_i(P_{split})
    \label{eq:cei_standard}
\end{equation}

\begin{equation}
    \textrm{CEI}_\mathrm{E} (P_{split}) = 1-\frac{1}{\log_2K}\max_{i}(S^{\prime}_i(P_{split}))
    \label{eq:cei_extreme}
\end{equation}

\noindent Both CEI\textsubscript{N} and CEI\textsubscript{E} range from $0$ to $1$, where higher values indicate a fairer model. In our original work~\cite{2024_cei_icpr}, the CEI\textsubscript{E} variant, configured with $P_{split}$ at the $95^{th}$ percentile and weights $(w_t, w_c) = (0.8, 0.2)$, was found to provide highly representative scores, and we utilize this configuration in parts of our experimentation.

\subsubsection{Automated Parameter Selection for CEI (CEI\textsuperscript{A})}
\label{subsubsec:CEI_auto_params}

The original CEI formulation relies on manually chosen hyperparameters: the percentile $P_{split}$ to divide the distributions and the weights $w_t$ and $w_c$. Although this offers flexibility, it can also require additional configuration steps. To simplify this process and enhance objectivity, we propose an automated method for selecting these parameters, resulting in a version we term CEI\textsuperscript{A}. To determine a meaningful percentile $P_{split}$ for dividing each distribution, we employ a statistical approach based on the N-sigma rule. For a given score distribution with mean $\mu$ and standard deviation $\sigma$, we first calculate a score threshold $t$ based on a chosen number of standard deviations, $N$:
\begin{equation}
    t = \mu + N \cdot \sigma \quad (\text{for right tail, or } t = \mu - N \cdot \sigma \text{ for left tail})
    \label{eq:cei_threshold_auto}
\end{equation}
Once this threshold $t$ is established, the empirical percentile $P(t)$ corresponding to it is determined by calculating the proportion of data points in the distribution that are less than or equal to $t$ (or greater than or equal to $t$, for a left-tail focus):
\begin{equation}
    P(t) = \left( \frac{1}{n} \sum_{j=1}^{n} \mathbf{1}_{\{x_j \leq t\}} \right) \cdot 100
    \label{eq:percentile_from_th}
\end{equation}
\noindent where $x_j$ are the scores in the distribution and $n$ is the total number of scores. This $P(t)$ then serves as our $P_{split}$. This method allows the split point to adapt to the dispersion of each distribution and ensures that the analysis focuses on regions of specified statistical extremity. The choice of $N$ (and thus $P_{split}$) can be customized to the desired sensitivity to rare events.

The weights $w_t$ and $w_c$ in Eq.~\ref{eq:s_prime} determine the emphasis placed on the tail versus the center of the distributions. To automate their selection, we aim to make $w_t$ reflect the "heaviness" or "abnormality" of each demographic group's distribution tail relative to a baseline, such as a normal distribution. For each demographic group $d_i$, we first quantify the deviation of its tail from this baseline. Let $\Delta_i$ be a precalculated measure representing this difference in probability mass or shape for the tail of the $d_i$' group distribution (e.g., how much more mass is in the empirical tail compared to the tail of a fitted normal distribution at the same $P_{split}$). We then compute an intermediate scalar, $s_i$, for each group:
\begin{equation}
    s_i = \frac{1}{1 + e^{-\Delta_i}} \cdot \frac{P(t)}{100}
    \label{eq:cei_weights_scalar_auto}
\end{equation}
\noindent Here, the sigmoid function $\frac{1}{1 + e^{-\Delta_i}}$ maps the tail deviation $\Delta_i$ to a value between $0$ and $1$. This value is then scaled by $P(t)/100$, where $P(t)$ is the automatically determined percentile (from Eq.~\ref{eq:percentile_from_th}). This scaling implies that the influence of the tail's "heaviness factor" ($s_i$) on the overall tail weight $w_t$ is proportional to the size of the region defined as the tail.

The final automated tail weight, $w_t^\mathrm{A}$, is the average of these individual $s_i$ values across all $K$ demographic groups:
\begin{equation}
    w_t^\mathrm{A} = \frac{1}{K}\sum_{i=1}^K s_i
    \label{eq:cei_weights_auto}
\end{equation}
The center weight, $w_c^\mathrm{A}$, is then simply the complement: $w_c^\mathrm{A} = 1 - w_t^\mathrm{A}$. These automated weights $w_t^\mathrm{A}$ and $w_c^\mathrm{A}$ are then used in Eq.~\ref{eq:s_prime} for the CEI\textsuperscript{A} calculation.

%% file: sections/4_experimental_protocol.tex
\section{Experimental Framework}\label{sec:experimental_framework}

This section details the experimental framework used to validate our proposed CEI metric. We assess demographic bias in face recognition models under various training conditions, data distributions (balanced to skewed), and model architectures, using both our own trained models and publicly available state-of-the-art systems.

\subsection{Datasets}
\label{subsubsec:datasets}

Our analysis employs diverse training and evaluation datasets, varying in scale, demographic composition, and realism, to investigate model behavior under different data conditions. The key properties of these data sets are summarized in Table~\ref{table:datasets_properties}.

\begin{table}[ht]
    \centering
    \caption{Specifications of the face recognition training and testing datasets used over the different experiments.}
    \label{table:datasets_properties}
    \begin{tabular}{llccc}
        \toprule
        \textbf{Dataset} & \textbf{Purpose} & \textbf{\#Identities} & \textbf{\#Images} & \textbf{Demographic Annotations}\\
        \midrule
        BUPT-B~\cite{wang2020mitigating} & Training & 28K & 1.3M & Yes\\
        MS1MV2~\cite{guo2016MS-Celeb} & Training & 87K & 5.8M & No \\
        MS1MV3~\cite{Deng2018ArcFaceAA} & Training & 93K & 5.2M & No \\
        Glint360K~\cite{an2021partial} & Training & 360K & 17.1M & No \\
        WebFace4M~\cite{Zhu2021WebFace260MAB} & Training & 200K & 4.3M & No \\
        WebFace42M~\cite{Zhu2021WebFace260MAB} & Training & 2M & 42.5M & No \\
        \midrule
        RFW~\cite{Wang2018RacialFI} & Testing & 12K & 40K & Yes \\
        MORPH~\cite{Ricanek2006MORPH} & Testing & 13.6K & 55.1K & Yes \\
        GANDiffFace~\cite{Melzi2023GANDiffFaceCG} & Testing & 10K & 500K & Yes \\
        \bottomrule
    \end{tabular}
\end{table}
\paragraph{Training Datasets.}
The following training datasets (detailed in Table~\ref{table:datasets_properties}) were used:

\noindent \textbf{- BUPT-B}~\cite{wang2020mitigating}: Used for its demographic labels, allowing controlled experiments that include balanced training and the intentional introduction of bias by training on specific demographic subgroups.

\noindent \textbf{- MS1MV3}~\cite{Deng2018ArcFaceAA}, and \textbf{Glint360K}~\cite{an2021partial}: These large-scale datasets, often unlabeled in terms of demographic data, and used to pretrain public models (e.g., from InsightFace\footnote{\url{https://github.com/deepinsight/insightface/tree/master/model_zoo#1-face-recognition-models}}, represent typical data for developing high-performance systems.

\noindent \textbf{- WebFace4M and WebFace42M}~\cite{Zhu2021WebFace260MAB}: Subsets of the WebFace260M benchmark, used to assess how bias manifests with varying large data scales.

\paragraph{Evaluation Datasets.}
For evaluation, we selected datasets with available demographic annotations (see Table~\ref{table:datasets_properties}):

\noindent \textbf{- RFW}~\cite{Wang2018RacialFI}: Used for racial bias analysis with images captured unconstrained "in the wild" \cite{2018_TIFS_SoftWildAnno_Sosa}.

\noindent \textbf{- MORPH}~\cite{Ricanek2006MORPH}: Contains high-quality \cite{2022_WACVw_FaceQvec_JHO} controlled images with annotations for race, gender, and age, facilitating multi-attribute disparity analysis.

\noindent \textbf{- GANDiffFace}~\cite{Melzi2023GANDiffFaceCG}: A dataset of synthetic images with controlled facial attributes, which allows systematic fairness evaluations under well-defined conditions and relevant to recent challenges (e.g., FRCSyn~\cite{melzi2024frcsyn}).

\subsection{Face Recognition Models}
\label{subsubsec:biometric_models}

We evaluated three model categories: models trained in-house for controlled studies, publicly available pre-trained models (e.g., from InsightFace), and other high-performance baseline models.

\textit{In-house Trained Models}
We trained ResNet-50 models in BUPT-B using CosFace loss~\cite{Wang2018CosFaceLM}: one on the full dataset (the 'All Groups' model) and others intentionally biased by training on single demographic groups. Furthermore, ResNet-50 and ResNet-100 models were trained on WebFace4M for experiments where a finer-grained training control was not required.

\textit{Publicly Available Models}
We include the ResNet-50 and ResNet-100 models from InsightFace, pre-trained with ArcFace loss~\cite{Deng2018ArcFaceAA} on large-scale datasets (MS1MV3, Glint360K). These serve as reference points for bias evaluation in widely used models.

\textit{High-performance Baselines}
To contextualize fairness against accuracy, we include two strong benchmark models: a ResNet-100 trained on WebFace42M and a proprietary model. Both demonstrate high performance on standard benchmarks like IJB-C~\cite{Maze2018158IJBC} (e.g. ResNet-100: FNMR $0.0407$ @ FMR $10^{-4}$; proprietary: FNMR $0.037$ @ FMR $10^{-4}$ and FNMR $0.0058$ @ FMR $0.0003$ on NIST FRTE $1:1$).

%% file: sections/5_results_and_interpretation.tex
\section{Experiments: Results and Interpretation}\label{sec:experiments_and_interpretation}

Face recognition (FR) models, trained on data, can inadvertently learn and embed biases. Evaluating these biases in operational "black-box" systems is challenging due to limited access to internal model details or original training data. Typically, only output similarity scores are available, underscoring the need for robust output-based metrics like our proposed Comprehensive Equity Index (CEI).

To validate CEI and investigate the factors that influence demographic bias (e.g., composition of training data, model architecture, scale of data set), we conducted four experiments, using the resources detailed in Sect.~\ref{sec:experimental_framework}. These four experiments include: i) evaluating CEI using synthetically-generated distributions to test its sensitivity and behavior in precisely controlled bias scenarios; ii) assessing CEI's performance against existing metrics in operational biometric systems on real-world datasets; iii) analyzing CEI in models intentionally trained with varying levels of bias; and iv) applying CEI to high-performance models trained on large, uncontrolled web-collected data to investigate bias.

\begin{figure*}[!h]
    \centering
    \includegraphics[width=\textwidth]{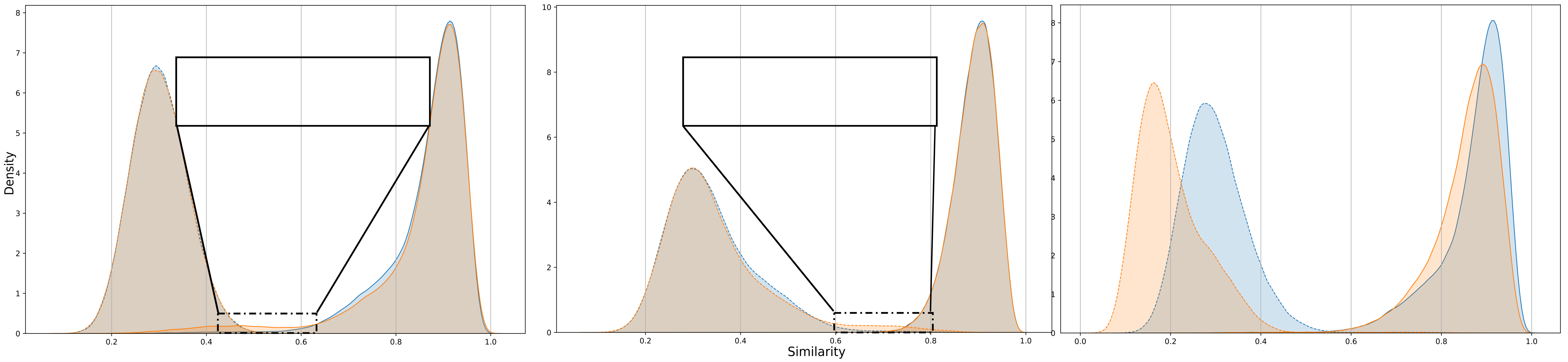}
    \caption{Genuine and impostor synthetically-generated similarity score distributions, in different scenarios: (Left) Biased Genuine distribution tail (BG); (Center) Biased Impostor distribution tail (BI); and (Right) Biased genuine-impostor distribution Center (BC).}
    \label{fig:syntheticcases2}
\end{figure*}

\subsection{Synthetically-Generated Distributions}
\label{subsubsec:exp_synthetic_dis}

To assess bias estimation metrics under controlled conditions, we performed synthetic experiments using Beta-distributed~\cite{johnson1995beta} genuine and impostor scores, simulating a high-performance biometric system. These simulations allow precise manipulation of score distribution characteristics (tail, shape, center) to analyze metric sensitivity to various distributional shifts.
We defined three scenarios (Figure~\ref{fig:syntheticcases2}): \textbf{Biased Genuine distribution tail (BG)}, where a longer lower tail in genuine scores increases false rejections; \textbf{Biased Impostor distribution tail (BI)}, where a longer upper tail in impostor scores increases false acceptances; and \textbf{Biased genuine-impostor distribution Centers (BC)}, where distribution means are shifted while tail behavior is largely preserved, simulating systematic deviations not necessarily in the tails.
This experiment aims to: 1) understand the behavior of our automated parameter selection for CEI (Eq.~\ref{eq:percentile_from_th}, Eq.~\ref{eq:cei_weights_auto}), and 2) identify which metrics detect the introduced biases. We hypothesize that tail-sensitive metrics (GARBE\textsubscript{FMR}, GARBE\textsubscript{FNMR}, IN\textsubscript{FMR}, and IN\textsubscript{FNMR}) will respond in the BG/BI scenarios, while distribution-shaped metrics (DFI\textsubscript{N} and DFI\textsubscript{E}) will be more sensitive in the BC scenario.

\subsubsection{Analyzing the behavior of automatized parameters in CEI\textsuperscript{A}}
\label{subsubsec:synthetic_cei_parameters}

Equations~\ref{eq:percentile_from_th} and~\ref{eq:cei_weights_auto} define the automated selection of percentiles and tail weights for CEI, respectively, based on \(N\), the number of standard deviations from the mean. Percentile selection generally stabilizes or maximizes effectiveness around $N=1-2$ sigmas, obtaining the $84th$ and $94th$ percentile, respectively. Similarly, for tail weight \(w_{tail}\), \(N=2\) sigmas generally produce high tail relevance (\(w_{tail} \approx 0.71\)), although impostor distributions in cases of BI and BG benefit from \(N=3\) sigmas ($98th$ percentile). For simplicity in this study, we adopt \(N=3\) sigmas to calculate the automated CEI version (CEI\textsuperscript{A}).

\subsubsection{Obtained results under a synthetic use case}
\label{subsubsec:synthetic_metrics_results}

\begin{table}[t!]
\centering
\caption{DFI, NIST metrics, and the proposed CEI\textsubscript{E} and CEI\textsuperscript{A} on the synthetic scenario. *E refers to the extreme variant, while G and I refer to the Genuine and Impostor distributions, respectively.}
\label{table:sota_metrics_synthetic}
    \begin{tabular}{lccc}
    \hline
    \multicolumn{1}{l}{\textbf{Metrics}} & \textbf{BG} & \textbf{BI} & \textbf{BC} \\
    \hline
    DFI\textsubscript{N} & 0.9983 & 0.9974 & 0.8361 \\
    DFI\textsubscript{E} & 0.9982 & 0.9970 & 0.8112 \\
    \hline
    GARBE\textsubscript{FMR} & 0.0050 & 0.2950 & 0.0433 \\
    GARBE\textsubscript{FNMR} & 0.3326 & 0.0025 & 0.0208 \\
    \hline
    IN\textsubscript{FMR} & 1.1249 & 2.0989 & 1.0697 \\
    IN\textsubscript{FNMR} & 2.2331 & 1.0037 & 1.0416 \\
    \hline
    CEI\textsubscript{E,G}* & 0.4714 & 0.9990 & 0.9916 \\
    CEI\textsubscript{E,I}* & 0.9992 & 0.5372 & 0.2986 \\
    \hline
    CEI\textsuperscript{A}\textsubscript{N,G}* & 0.5446 & 0.9968 & 0.9968 \\
    CEI\textsuperscript{A}\textsubscript{N,I}* & 0.9831 & 0.5666 & 0.5654 \\
    CEI\textsuperscript{A}\textsubscript{E,G}* & 0.4554 & 0.9967 & 0.9967 \\
    CEI\textsuperscript{A}\textsubscript{E,I}* & 0.9783 & 0.4751 & 0.4739 \\
    \hline
    \end{tabular}
\end{table}

Table~\ref{table:sota_metrics_synthetic} presents the metric performances in this controlled setting. As hypothesized, DFI\textsubscript{N} and DFI\textsubscript{E} showed limited sensitivity to tail biases in the BG and BI scenarios (indicated by high similarity scores, suggesting low bias detection), confirming their potential insufficiency for such subtle biases. In contrast, GARBE and IN metrics effectively detected these tail manipulations, highlighting their utility when bias manifests at extreme operating points.

In the BC scenario, where the distribution centers were shifted but the tails remained similar, the DFI metrics accurately identified bias, while the tail-focused outcome-based metrics (GARBE, IN) showed less responsiveness. These findings confirm that different types of metrics offer complementary insight into bias, and no single metric fully characterizes all aspects. In particular, both the original CEI formulation and its proposed automated variant (CEI\textsuperscript{A}) provide an additional layer of interpretability by clearly distinguishing whether the genuine or impostor distribution contributes more significantly to the detected bias.

\subsection{Evaluation in Operational Biometric Systems}\label{subsubsec:exp_real}

\begin{figure*}[!h]
    \centering
    \includegraphics[width=\textwidth]{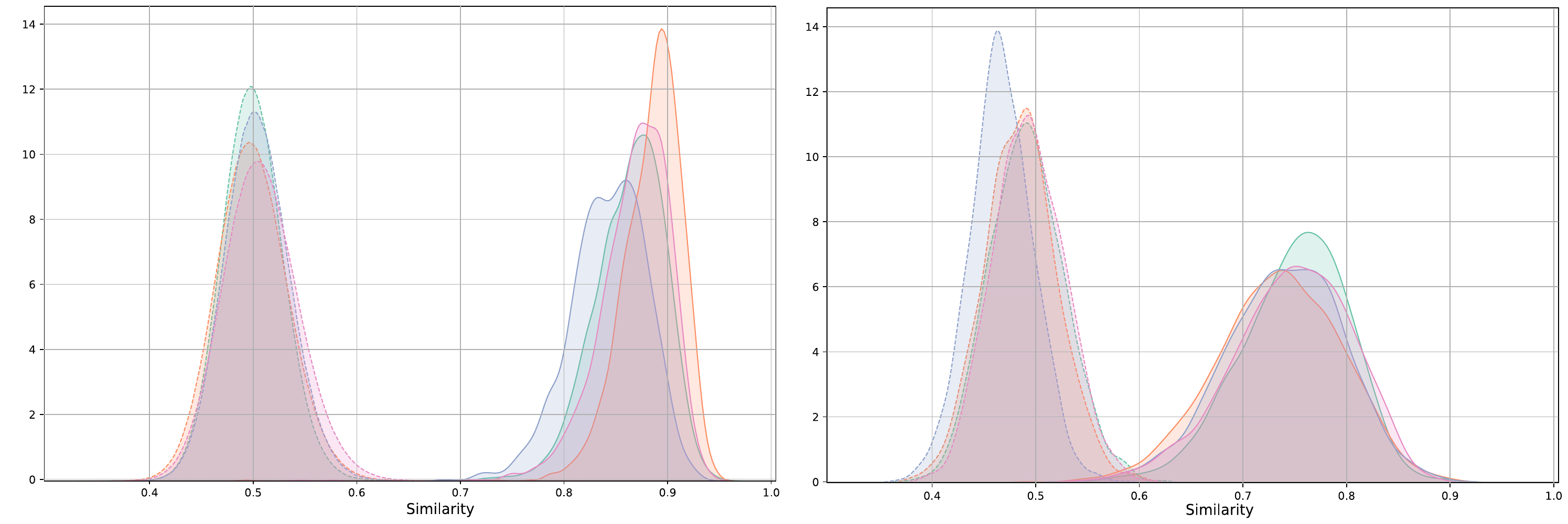}
    \caption{Genuine (continuous line) and impostor (dashed line) distributions for ResNet-100~\cite{He2015DeepRL} model in MORPH~\cite{Ricanek2006MORPH} (Left) and RFW~\cite{Wang2018RacialFI} (Right) datasets. The x-axis shows the cosine similarity between two images. Thus the impostor distributions are on the left and the genuine on the right. Each demographic group is represented by a different color.}
    \label{fig:real_cases_r100}
\end{figure*}

Figure~\ref{fig:real_cases_r100} displays the genuine and impostor score distributions for a ResNet-100 model~\cite{He2015DeepRL} (trained on WebFace42M~\cite{Zhu2021WebFace260MAB}) when evaluated on the MORPH~\cite{Ricanek2006MORPH} and RFW~\cite{Wang2018RacialFI} datasets, showing variations between ethnic groups. This analysis focuses on our improved and automated CEI variant, CEI\textsuperscript{A}, particularly its extreme version (CEI\textsuperscript{A}\textsubscript{E}). This version automatically determines percentiles and tail weights ($w\textsubscript{tail}$, with $w\textsubscript{no\_tail} = 1 - w\textsubscript{tail}$) using a specified $N$-sigma value, as per Eq.~\ref{eq:percentile_from_th} and Eq.~\ref{eq:cei_weights_auto}.

Table~\ref{table:cei_auto_real_config} details these automatically derived parameters and the resulting CEI\textsuperscript{A}\textsubscript{E} scores for $N \in \{1, 2, 3\}$ with the ResNet-100 model. Consistent with synthetic experiments, $N=2$ achieves high percentiles ($>95\%$), often reaching $\approx 99\%$ by $N=3$, and the tail weight $w\textsubscript{tail}$ increases with $N$. In fact, $w\textsubscript{tail}$ generally stabilizes around $0.7$. This value effectively balances tail sensitivity with information from the distribution's center. Based on these observations, we select $N=3$ for subsequent CEI\textsuperscript{A} evaluations to ensure robust tail emphasis.

\begin{table}[t!]
    \caption{Study of the automatic CEI version. We show the obtained configuration (percentile and tail weight) values, and the final CEI\textsubscript{E} metric score for three different values for \(N\)-sigma ($1$, $2$, $3$). We are using the ResNet-100~\cite{He2015DeepRL} model on RFW~\cite{Wang2018RacialFI}, and MORPH~\cite{Ricanek2006MORPH}.}
    \label{table:cei_auto_real_config}
    \centering
    \setlength{\tabcolsep}{3pt}
    \resizebox{\textwidth}{!}{
    \begin{tabular}{clccccccc}
        \toprule
        & & \multicolumn{3}{c}{RFW} & \multicolumn{3}{c}{MORPH} \\
        \cmidrule(lr){3-5} \cmidrule(lr){6-8}
        \(N-\)sigma & Distribution & Percentile & Tail Weight & Metric Score & Percentile & Tail Weight & Metric Score \\
        \midrule
        \multirow{2}{*}{$1$} 
        & Genuines & 84 & $0.61$ & $0.928$ & 85 & $0.62$ & $0.722$ \\
        & Impostors & 84 & $0.61$ & $0.791$ & 84 & $0.61$ & $0.940$ \\
        \midrule
        \multirow{2}{*}{$2$} 
        & Genuines & 98 & $0.71$ & $0.881$ & 99 & $0.73$ & $0.579$ \\
        & Impostors & 97 & $0.72$ & $0.729$ & 97 & $0.71$ & $0.843$ \\
        \midrule
        \multirow{2}{*}{$3$} 
        & Genuines & 99 & $0.72$ & $0.848$ & 99 & $0.69$ & $0.596$ \\
        & Impostors & 99 & $0.71$ & $0.712$ & 999 & $0.72$ & $0.814$ \\
        \bottomrule
    \end{tabular}
    }
\end{table}

Table~\ref{table:sota_metrics_side_by_side} presents a comparative evaluation of our CEI proposal (CEI\textsubscript{E}, CEI\textsuperscript{A}\textsubscript{N}, and CEI\textsuperscript{A}\textsubscript{E}, using $N=3$) against DFI and NIST-related metrics in various operational datasets. Consistent with the hypotheses, the DFI\textsubscript{N} and DFI\textsubscript{E} metrics do not capture the demographic differences evident in these systems. This is likely due to their aggregation of genuine and impostor distributions and lower sensitivity to discrepancies primarily located in the distribution tails. In contrast, NIST-related metrics (GARBE, IN), by assessing genuine and impostor distributions separately, can detect differences between demographic groups and indicate whether these originate from genuine or impostor score behaviors, offering more granular insights. Our proposed CEI metrics (both the manual CEI\textsubscript{E} and the automatized CEI\textsuperscript{A} variants) significantly outperform DFI in identifying the disparities that DFI overlooks. In particular, the extreme variants (CEI\textsubscript{E} and CEI\textsuperscript{A}\textsubscript{E}) exhibit greater sensitivity to quantifying these demographic differences. These findings validate the proposed CEI versions as effective tools that capture existing demographic differences, similar to differential outcome-based metrics, while retaining the strengths of a performance-based fairness assessment approach. From now on, we will stick to the extreme variant in the proposed CEI\textsuperscript{A} metric as well. Further observations in Table~\ref{table:sota_metrics_side_by_side} reveal that CEI\textsuperscript{A}\textsubscript{E} scores often differ significantly for genuine and impostor distributions within the same data set. This indicates distinct bias characteristics for false matches versus false non-matches, with these metric scores highlighting disparities across demographic groups that are visually supported by the score distributions in Fig.~\ref{fig:real_cases_r100}.

\begin{table}[t!]
    \caption{Values of the DFI and NIST-related metrics, and the proposed CEI versions on operational systems, evaluated on MORPH~\cite{Ricanek2006MORPH}, RFW~\cite{Wang2018RacialFI}, and GANDiffFace~\cite{Melzi2023GANDiffFaceCG}.}
    \label{table:sota_metrics_side_by_side}
    \centering
    \scriptsize
    \setlength{\tabcolsep}{3pt}
    \begin{tabular}{@{}lcccccc@{}}
        \toprule
        \textbf{Metric} & \multicolumn{2}{c}{\textbf{MORPH}} & \multicolumn{2}{c}{\textbf{RFW}} & \multicolumn{2}{c}{\textbf{GANDiffFace}} \\
        \cmidrule(lr){2-3} \cmidrule(lr){4-5} \cmidrule(lr){6-7}
        & \textbf{ResNet-100} & \textbf{Proprietary} & \textbf{ResNet-100} & \textbf{Proprietary} & \textbf{ResNet-100} & \textbf{Proprietary} \\
        \midrule
        DFI\textsubscript{N} & $0.9932$ & $0.9933$ & $0.9785$ & $0.9818$ & $0.9927$ & $0.9906$ \\
        DFI\textsubscript{E} & $0.9885$ & $0.9873$ & $0.9529$ & $0.9647$ & $0.9768$ & $0.9662$ \\
        \midrule
        GARBE\textsubscript{FMR} & $0.3762$ & $0.2439$ & $0.2885$ & $0.2500$ & $0.4631$ & $0.4616$ \\
        GARBE\textsubscript{FNMR} & $0.1500$ & $0.1500$ & $0.1377$ & $0.1941$ & $0.0654$ & $0.0693$ \\
        \midrule
        IN\textsubscript{FMR} & $2.9418$ & $2.9410$ & $1.8803$ & $1.7965$ & $4.3461$ & $4.1876$ \\
        IN\textsubscript{FNMR} & $1.6818$ & $1.6818$ & $1.3723$ & $1.5635$ & $1.1754$ & $1.2038$ \\
        \midrule
        CEI\textsubscript{E,G} & $0.5425$ & $0.7953$ & $0.6724$ & $0.6408$ & $0.6717$ & $0.6704$ \\
        CEI\textsubscript{E,I} & $0.6797$ & $0.8585$ & $0.3940$ & $0.4560$ & $0.6989$ & $0.6867$ \\
        \midrule
        CEI\textsuperscript{A}\textsubscript{N,G}* & $0.5958$ & $0.5384$ & $0.8480$ & $0.7989$ & $0.8437$ & $0.8056$ \\
        CEI\textsuperscript{A}\textsubscript{N,I}* & $0.8141$ & $0.9271$ & $0.7121$ & $0.7457$ & $0.8410$ & $0.8307$ \\
        CEI\textsuperscript{A}\textsubscript{E,G}* & $0.3603$ & $0.3283$ & $0.7905$ & $0.7522$ & $0.5366$ & $0.6435$ \\
        CEI\textsuperscript{A}\textsubscript{E,I}* & $0.6216$ & $0.8550$ & $0.4586$ & $0.5523$ & $0.7182$ & $0.5887$ \\
        \bottomrule
    \end{tabular}
    \begin{flushleft}
        \footnotesize 
        The manual CEI variants use a percentile of $95\%$ and weights ($w\textsubscript{tail}, w\textsubscript{center})=(0.8, 0.2)$. CEI\textsuperscript{A} variants use N=3. *E refers to the extreme variant, while G and I refers to the Genuine and Impostor distribution, respectively.
    \end{flushleft}
\end{table}

\subsection{Evaluation in Controlled-Bias Scenarios}
\label{subsec:exp_buptb_biased}

To further assess CEI's ability to detect and quantify demographic bias, we intentionally trained models biased towards specific demographic groups using the BUPT-B dataset~\cite{wang2020mitigating}. BUPT-B is suitable due to its balanced demographic coverage (four major ethnic groups) and allowance for controlled training data composition.

We trained five ResNet-50~\cite{He2015DeepRL} models with identical protocols: a reference model on the complete BUPT-B dataset (referred to as 'All Groups') and four models each trained exclusively on data from a single ethnic group to induce specific demographic biases. These single-group models were designed to induce specific demographic biases relative to the other groups. All models were evaluated on the RFW benchmark~\cite{Wang2018RacialFI} for cross-demographic performance analysis.

\begin{table}[t!]
\caption{Bias metrics on RFW for models biased using BUPT-B training. DFI metrics~\cite{kotwal2023fairness}, GARBE metrics~\cite{howard2019effect}, IN metrics~\cite{grother2022face}. *E refers to the extreme variant, while G and I refer to the Genuine and Impostor distribution, respectively.}
\label{table:bupt_b_biased_training}
    \centering
    \setlength{\tabcolsep}{3pt}
    \begin{tabular}{@{}lccccc@{}}
    \toprule
    Metric & All & Grp. 1 & Grp. 2 & Grp. 3 & Grp. 4 \\ %
    \midrule
    DFI\textsubscript{N} & $0.9849$ & $0.9890$ & $0.9887$ & $0.9888$ & $0.9842$ \\
    DFI\textsubscript{E} & $0.9802$ & $0.9813$ & $0.9817$ & $0.9794$ & $0.9801$ \\
    \midrule
    GARBE\textsubscript{FMR} & $0.4582$ & $0.4806$ & $0.5177$ & $0.5191$ & $0.5498$ \\
    GARBE\textsubscript{FNMR} & $0.0856$ & $0.0260$ & $0.0289$ & $0.0205$ & $0.1164$ \\
    \midrule
    IN\textsubscript{FMR} & $3.9980$ & $4.3661$ & $5.3810$ & $5.6192$ & $6.4426$ \\
    IN\textsubscript{FNMR} & $1.2500$ & $1.0484$ & $1.0658$ & $1.0443$ & $1.3561$ \\
    \midrule
    CEI\textsubscript{E,G}* & $0.7516$ & $0.8032$ & $0.8087$ & $0.8230$ & $0.6955$ \\
    CEI\textsubscript{E,I}* & $0.5959$ & $0.5217$ & $0.5559$ & $0.5443$ & $0.5716$ \\
    \midrule
    CEI\textsuperscript{A}\textsubscript{E,G}* & $0.6709$ & $0.8498$ & $0.7295$ & $0.6179$ & $0.6949$ \\
    CEI\textsuperscript{A}\textsubscript{E,I}* & $0.5039$ & $0.4446$ & $0.4822$ & $0.4390$ & $0.4577$ \\
    \bottomrule
    \end{tabular}
\end{table}

Table~\ref{table:bupt_b_biased_training} summarizes the bias assessments of DFI, IN, GARBE, and our CEI metric for these models. Consistent with previous findings, DFI metrics showed limited sensitivity to demographic differences in these controlled bias settings. Other metrics indicated more pronounced differences in impostor distributions, which can impact system security (false matches). Specifically, IN\textsubscript{FMR} and GARBE\textsubscript{FMR} values increased significantly for the single-group biased models (Groups 2, 3, and 4), successfully reflecting the expected higher FMR disparities when a model is not trained comprehensively. Among the CEI\textsuperscript{A} variants, the extreme version (CEI\textsuperscript{A}\textsubscript{E}) consistently captured more significant differences.

Regarding genuine distributions, metrics suggested that training on the balanced BUPT-B data set 'All Groups' did not necessarily minimize bias; this model sometimes exhibited higher bias scores than certain single-group trained models (e.g., the Group 1 biased model). This non-uniform bias, despite balanced demographic training samples, highlights that fairness is not guaranteed by simple count-balancing alone. This aligns with observations by Wang \textit{et al.}~\cite{wang2020mitigating} using BUPT-B. It underscores that while data balance is crucial, other factors such as image acquisition quality~\cite{terhorst2022comprehensive} also contribute to biases, which are not due solely to demographic under-representation.

\subsection{Evaluation in Uncontrolled Biased Trainings}\label{subsec:webface4m}

This experiment evaluates models trained on large-scale, web-collected datasets commonly used for high-performance FR: MS1MV3~\cite{Deng2018ArcFaceAA}, Glint360K~\cite{an2021partial}, and WebFace4M~\cite{Zhu2021WebFace260MAB}. These datasets typically lack detailed demographic annotations, making fairness evaluation relying on downstream tests. This setup allows us to study bias in models trained on massive, unfiltered data and explore whether such data mitigates or reinforces existing disparities.

We evaluated in-house trained ResNet-50 and ResNet-100~\cite{He2015DeepRL} models (on WebFace4M) alongside publicly available InsightFace models (ResNet architectures on MS1MV3 and Glint360K) on the RFW dataset~\cite{Wang2018RacialFI} (Table~\ref{table:uncontrolled_performance_rfw_no_error_restructured}) to reveal performance variations between demographic groups, with the Caucasian group often achieving the best results. In particular, significant performance drops were observed for non-Caucasian groups with certain models.

\begin{table}[t!]
\caption{Performance in RFW for models trained under uncontrolled conditions measured in TPR@FPR=$10^{-3}$.}
\label{table:uncontrolled_performance_rfw_no_error_restructured}
    \centering
    \small
    \resizebox{\columnwidth}{!}{\begin{tabular}{@{}llcccc@{}}
        \toprule
        Dataset & Model & African & Asian & Caucasian & Indian \\
        \midrule
        \multirow{2}{*}{WebFace4M} & ResNet50  & $0.8757$ & $0.8710$ & $0.9697$ & $0.9097$ \\
                                   & ResNet100 & $0.9357$ & $0.9350$ & $0.9843$ & $0.9407$ \\
        \midrule
        \multirow{2}{*}{MS1MV3}    & ResNet50  & $0.9597$ & $0.9577$ & $0.9903$ & $0.9757$ \\
                                   & ResNet100 & $0.9850$ & $0.9613$ & $0.9930$ & $0.9847$ \\
        \midrule
        \multirow{2}{*}{Glint360K} & ResNet50  & $0.9613$ & $0.9363$ & $0.9877$ & $0.9620$ \\
                                   & ResNet100 & $0.9650$ & $0.9473$ & $0.9900$ & $0.9743$ \\
        \bottomrule
    \end{tabular}}
\end{table}

Table~\ref{table:bias_metrics_uncontrolled} presents the bias metrics for these models, using $N=3$ for CEI\textsuperscript{A} (yielding the $99^{th}$ percentile and $w_{tail} \approx 0.7$), consistent with previous experiments.

\begin{table}[t!]
\centering
\caption{Bias metrics on RFW for models (ResNet-50 and ResNet-100) trained on different datasets. DFI metrics from Kotwal et al.~\cite{kotwal2023fairness}, GARBE metrics from Howard et al.~\cite{howard2019effect}, IN metrics from Grother et al.~\cite{grother2022face}. *E refers to the extreme variant, while G and I refer to the Genuine and Impostor distribution, respectively.}
\label{table:bias_metrics_uncontrolled}
    \small
    \resizebox{\columnwidth}{!}{
    \begin{tabular}{@{}lcccccc@{}} 
        \toprule
        & \multicolumn{2}{c}{WebFace4M} & \multicolumn{2}{c}{MS1MV3} & \multicolumn{2}{c}{Glint360K} \\
        \cmidrule(lr){2-3} \cmidrule(lr){4-5} \cmidrule(lr){6-7}
        Metric & R50 & R100 & R50 & R100 & R50 & R100 \\
        \midrule
        DFI\textsubscript{N} & $0.9776$ & $0.9805$ & $0.9799$ & $0.9802$ & $0.9740$ & $0.9745$ \\
        DFI\textsubscript{E} & $0.9475$ & $0.9574$ & $0.9519$ & $0.9512$ & $0.9355$ & $0.9354$ \\
        \midrule
        GARBE\textsubscript{FMR} & $0.3748$ & $0.4037$ & $0.5191$ & $0.3332$ & $0.3331$ & $0.2915$ \\
        GARBE\textsubscript{FNMR} & $0.1412$ & $0.1366$ & $0.0785$ & $0.1412$ & $0.1322$ & $0.1260$ \\
        \midrule
        IN\textsubscript{FMR} & $2.7088$ & $3.0408$ & $5.6192$ & $2.2350$ & $2.4483$ & $1.9872$ \\
        IN\textsubscript{FNMR} & $1.4644$ & $1.4367$ & $1.2543$ & $1.4207$ & $1.4666$ & $1.3676$ \\
        \midrule
        CEI\textsubscript{E,G} & $0.8075$ & $0.8041$ & $0.8599$ & $0.8444$ & $0.8011$ & $0.8437$ \\
        CEI\textsubscript{E,I} & $0.3823$ & $0.3932$ & $0.3969$ & $0.3451$ & $0.4200$ & $0.3734$ \\
        \midrule
        CEI\textsuperscript{A}\textsubscript{E,G}* & $0.7018$ & $0.8176$  & $0.7924$ & $0.6559$ & $0.8126$ & $0.6555$ \\
        CEI\textsuperscript{A}\textsubscript{E,I}* & $0.3349$ & $0.3901$ & $0.3667$ & $0.3728$ & $0.3607$ & $0.4744$ \\
        \bottomrule
    \end{tabular}}
\end{table}

The extreme variant of CEI\textsuperscript{A} (CEI\textsuperscript{A}\textsubscript{E}) again tends to highlight more differences. Interestingly, the ResNet-100 model trained in MS1MV3, which showed high overall performance and relatively smaller performance gaps between RFW groups (Table~\ref{table:uncontrolled_performance_rfw_no_error_restructured}), does not uniformly exhibit the lowest bias according to all metrics. For example, impostor-focused metrics such as IN\textsubscript{FMR} and our CEI\textsubscript{E,Impostor}/CEI\textsuperscript{A}\textsubscript{E,Impostor} suggest that other models (e.g., ResNet-100 on Glint360K) might display less bias with respect to false matches.
Furthermore, while deeper architectures (ResNet-100 vs. ResNet-50 in the same training set) generally improved performance, particularly for non-Caucasian groups (as seen in Table~\ref{table:uncontrolled_performance_rfw_no_error_restructured}), the bias metrics in Table~\ref{table:bias_metrics_uncontrolled} often indicate that these performance gains are reflected in changes to the impostor distribution.

%% file: sections/6_conclusions.tex
\section{Conclusion}\label{sec:conclusions}

This work advanced demographic fairness measurement in biometric systems by introducing the Comprehensive Equity Index (CEI), a differential performance-based metric. CEI addresses limitations of previous methods by focusing on score distribution tails -where differences in high-performance Face Recognition (FR) models often reside- and enabling separate analysis of genuine and impostor distributions. Extensive experiments, including evaluations on state-of-the-art FR systems (detailed in Table \ref{table:sota_metrics_side_by_side}), intentionally biased models (Table \ref{table:bupt_b_biased_training}), and models from large-scale datasets (Table \ref{table:bias_metrics_uncontrolled}), demonstrated CEI's ability to capture subtle cross-group distributional differences.

A key contribution is CEI\textsuperscript{A}, an automated CEI version that algorithmically sets optimal percentile splits and tail weighting parameters, simplifying the application and improving objectivity. CEI\textsuperscript{A} effectively highlighted demographic disparities without manual adjustment. Configuring CEI\textsuperscript{A} with $N > 1$ sigma often resulted in high percentiles (e.g. $>95^{th}$), pinpointing significant differences between groups in extreme distribution tails. Automated weighting ensures that this focus in the tail is balanced with information from the distribution centers.

CEI, in both manual and automated (CEI\textsuperscript{A}) forms, overcomes noted deficiencies of prior differential performance metrics in operational settings. In evaluations on real-world models, CEI\textsuperscript{A} successfully quantified the visually apparent score disparities between groups (Fig.~\ref{fig:real_cases_r100}), offering more granular insight than single-value metrics by isolating genuine and impostor score biases, as shown in Sect.~\ref{subsubsec:exp_real}. Furthermore, our controlled-bias experiments highlighted a critical system limitation: performance disparities persist even when training data is demographically balanced, reinforcing that fairness is not solely a matter of data counts (Sect.~\ref{subsec:exp_buptb_biased}). Importantly, CEI retains the core advantages of differential performance methods, being threshold-independent and offering insights into intrinsic system behavior regarding demographic attributes.

CEI is a valuable complementary tool for fairness assessment. Although it detects distributional disparities, the definition of a system as ``fair" or ``biased" depends on the specific definition of fairness and the application context~\cite{2022_AI_SensitiveLoss_IS}. We recommend using CEI alongside traditional performance metrics (e.g., FMR, FNMR) and outcome-based differential indices for a complete understanding. A limitation of output-level metrics, including CEI, is their difficulty in fully quantifying bias without analyzing internal model representations. However, CEI can complement such embedding space studies~\cite{2022_PR_SetMargin_Morales}, as score distributions often reflect latent space characteristics. In this regard, the proposed CEI measures can be useful in combination to other existing approaches to mitigate bias~\cite{2022_AI_SensitiveLoss_IS}, and improving the invariance of feature representations across data populations~\cite{pena2021emotion}.

Future work includes exploring additional multilevel bias detection and evaluation. A promising direction is to use CEI to guide the training of FR systems, leveraging its ability to capture score distribution differences to potentially improve the fairness of embedding space representations~\cite{2022_PR_SetMargin_Morales}, possibly in conjunction with embedding-level analysis~\cite{serna25unravel}. Regarding our experimental setup, we used 9 face datasets of varying size and properties representing common setups in the literature. Anyway, there are other valuable datasets like FairFace~\cite{Karkkainen_2021_WACV}, and DiveFace~\cite{2022_AI_SensitiveLoss_IS} that should be considered in future work.

As stated in Sect..~\ref{sec:previous_work}, the developed methods are particularly useful for bias analysis across demographic groups in biometrics recognition, but are generally applicable to any problem related to comparing score tail distributions across different experimental conditions. In addition to demographic variations, in our future work, we will also study CEI methods in partial~\cite{2013PTomeFSI_FacialRegions,2022_INFFUS_Periocular_Alonso}, and occluded face recognition~\cite{sequeira25eccv}.

Last, our objective in this research line is to facilitate AI/ML explainability~\cite{ortega21xai,ortega23xai}, and auditability~\cite{alcala2025my,alcala25iccv} (specially in the area biometric systems). In this line we will try to exploit our experience in analyzing demographic variations in biometric systems to improve the auditability of LLMs and VLMs~\cite{mancera25mint-text,alcala25cvprw}.

\section{Acknowledgements}

This paper has been financed by the Government of Navarre within Industrial Doctorates 2022, the company Veridas\footnote{\url{https://veridas.com/en/}}, Cátedra ENIA UAM-VERIDAS en IA Responsible (NextGenerationEU PRTR TSI-100927-2023-2), project M2RAI (PID2024-160053OB-I00 MICIU/FEDER), and project BBforTAI (PID2021-127641OB-I00
MICINN/FEDER).